\documentclass{article}
\usepackage{spconf,amsmath,graphicx}
\usepackage{algorithm}
\usepackage{algorithmic}
\usepackage{xcolor}
\usepackage{amsfonts}
\newcommand\blfootnote[1]{%
  \begingroup
  \renewcommand\thefootnote{}\footnote{#1}%
  \addtocounter{footnote}{-1}%
  \endgroup
}

\usepackage{textcomp}

\usepackage{lipsum}

\title{RID-TWIN: AN END-TO-END PIPELINE FOR AUTOMATIC FACE DE-IDENTIFICATION IN VIDEOS}

\name{Anirban Mukherjee\thanks{Anirban Mukherjee and Monjoy Narayan Choudhury are funded by MINRO center at IIITB}, Monjoy Narayan Choudhury, Dinesh Babu Jayagopi}
\address{International Institute of Information Technology, Bangalore}
\begin{document}

\maketitle
\begin{abstract}

Face de-identification in videos is a challenging task in the domain of computer vision, primarily used in privacy-preserving applications. Despite the considerable progress achieved through generative vision models, there remain multiple challenges in the latest approaches. They lack a comprehensive discussion and evaluation of aspects such as realism, temporal coherence, and preservation of non-identifiable features. In our work, we propose RID-Twin: a novel pipeline that leverages the state-of-the-art generative models, and decouples identity from motion to perform automatic face de-identification in videos. We investigate the task from a holistic point of view and discuss how our approach addresses the pertinent existing challenges in this domain. We evaluate the performance of our methodology on the widely employed VoxCeleb2 dataset, and also a custom dataset designed to accommodate the limitations of certain behavioral variations absent in the VoxCeleb2 dataset. We discuss the implications and advantages of our work and suggest directions for future research.

\end{abstract}
\section{Introduction}
\label{sec:intro}

In this digital era, the preservation of identity is a crucial aspect of all publicly available content, especially in domains such as healthcare, video surveillance, social media, and bio-metrics. While the availability of human video datasets serves as a useful resource for several automation-related tasks such as activity recognition, human behavioral pattern recognition, and clinical analysis, it becomes a challenge to ensure that the identities of the subjects are not revealed \cite{ribaric2016identification}. There are various approaches to creating de-identified faces that are realistic in terms of human likeness and visual fidelity, and compatible with different tasks and scenarios. The earliest approaches relied on basic image-based editing methods such as pixelating or blurring \cite{harmon1973masking, harmon1973recognition}. With the help of generative vision models, it is possible to generate photo-realistic synthetic images. Deep learning based generative models such as Encoder-Decoder (\cite{qiu2022novel, gafni2019live}), GAN (\cite{sun2018hybrid, maximov2020ciagan, croft2019differentially}) and Diffusion  (\cite{uchida2022dedim, dockhorn2022differentially, jivri2023anonymizator}) have been used for de-identification tasks.  

\blfootnote{© 2024 IEEE. Personal use of this material is permitted. Permission from IEEE must be obtained for all other uses, in any current or future media, including reprinting/republishing this material for advertising or promotional purposes, creating new collective works, for resale or redistribution to servers or lists, or reuse of any copyrighted component of this work in other works.}

However, there are some challenges in the current methodologies in the context of automatic de-identification for videos. Most de-identification works are image-based \cite{hellmann2023ganonymization, meden2023face} approaches and cannot be applied to videos. Approaches that edit or inpaint the frames of videos individually might result in temporal inconsistencies. For face-swap-based approaches, it is challenging to find an appropriate actor-face that matches the source subject's facial features such as structure and ethnicity. Over the last few years, there has been a drastic development in the field of generative artificial intelligence (Generative AI) for tasks such as image synthesis, image captioning, and video generation. While they are used in various applications currently because of their highly realistic and diverse image generation quality, there is no video de-identification approach at the time of presenting this work that harnesses them effectively. Moreover, most of the de-identification approaches are evaluated based on only the quality of the output of the media but lack a domain-contextual quantitative evaluation and discussion on the preservation of features such as behaviors. These challenges cannot be overlooked in critical domains such as healthcare, where it is important to have a trustworthy system.

\begin{figure}[t]
    \includegraphics[width=8.5cm]{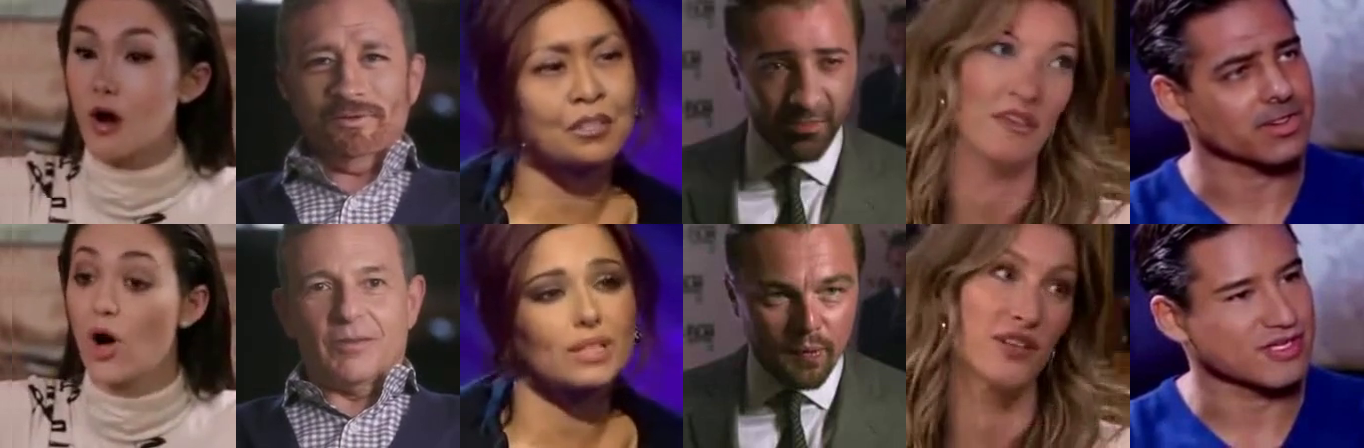}
    \caption{Sample outputs of RID-Twin: The images from source videos \textit{(bottom)} and their corresponding generated D-Twins \textit{(top)}}
    \label{fig:output_dtwins}
%
\end{figure}

In this context, our work addresses the following research questions: \textit{How can we build an effective pipeline for automatic de-identification of human faces in videos? What essential factors should it consider? How can we evaluate such a system?}

Addressing these questions, we propose a novel and effective pipeline as a viable solution. The key idea of our proposed model is to \textit{re-enact a de-identified twin actor}, thereby transferring motion to a de-identified image. Therefore, we call our methodology \textbf{RID-Twin}, which stands for \textbf{R}e-enacting \textbf{I}npainting based \textbf{D}e-identified-\textbf{Twin}. We utilize some of these best generative models to build a novel end-to-end automated pipeline. Fundamentally, its unique approach of disentangling motion from identity ensures an output that is both safe and realistic. We use evaluation metrics that can indicate how well a video has been de-identified and assess the pipeline's performance on the famous VoxCeleb2 dataset. Furthermore, we observe some of the difficult scenarios such as expression variations and extreme poses that are characteristic of real-world use cases but do not appear in the VoxCeleb2 dataset. To account for this absence, we propose a dataset encompassing such behavioral instances. In the end, we discuss the scope of our work, and how it addresses the pertinent challenges in this area, thus serving as a benchmark for future applications. In brief, the main contributions of our work are as follows:
\begin{itemize}
    \item We propose RID-Twin, a novel pipeline for fully automatic human de-identification of videos that uses the state-of-the-art generative models. We make the working code open source \footnote{https://github.com/AnirbanMukherjeeXD/RID-Twin} for public usage.
    \item We vigorously evaluate our pipeline with regards to the de-identification level, identity consistency, and expression preservation.
    
    \item We propose a dataset with a set of facial pose and expression variations, which could serve as a benchmark for future face de-identification tasks.
\end{itemize}

\section{Background}
\label{sec:background}
\subsection{Related works}

De-identification tasks have been approached broadly in the following three ways: Editing (introducing some transformations in the source face, mainly based on some parameters), Inpainting (filling the face area with a completely new generated face), and Face-swapping (using an actor's face as a proxy for the subject's face by swapping).

Editing-based methods ensure that the original structure of the face is preserved, but the identifiable aspects are changed. Some examples include works such as \cite{gafni2019live, meden2023face, croft2022differentially}. Successfully changing the attributes of the face while keeping the face realistic is a challenge. Too much alteration can result in an unrealistic or eerie-looking face, while too little alteration might leak the original identity. Some works such as \cite{meden2023face, dockhorn2022differentially, croft2022differentially}, in the lines of differential privacy, use a parameter $\epsilon$ to control the level of de-identification. In contrast, our approach deals with use cases where the output video must be completely devoid of any trace of identity in all situations. Most editing-based works de-identify at an image level. Ensuring uniform and consistent application of de-identification across all frames is a challenge for videos. To counter this issue, we chose not to use a frame-wise update approach.

Inpainting is also used in various de-identification tasks as shown in the works of \cite{li2019anonymousnet, balaji2021temporally, maximov2020ciagan, sun2018hybrid}. In these cases, the models successfully discard the existing face and generate a new one. This can be considered better than editing-based methods, as the de-identified face is a completely generated one, and not an edited version of the original one, as discussed earlier. However, preserving the original features across all the frames in a video is a similar challenge faced by editing-based approaches. As opposed to framewise inpainting, we chose to inpaint only one frame, and then subsequently animate that image for the remaining frames.

Face-swapping requires an actor-face, which is used to replace the subject's face. This can ensure that the identity is changed significantly since the key features of the de-identified faces are those of the actor's face. Works of \cite{thies2016face2face, yang2022face2face} perform face swapping and can be used as a tool for de-identification. While this idea is widely used for changing the identity of the actor, unsurprisingly the dependence on the actor's face itself poses a challenge. Since the swapped face should resemble the source face, finding an actor with a similar face/head structure, skin color, age group, wearing accessories and other features is a difficult task. To address this, we generate a synthetic actor that resembles our source subject in these non-identifiable aspects. In some state-of-the-art face-swapping models such as \cite{perov2020deepfacelab}, there is a need for manual post-processing tasks. This makes such approaches unsuitable for a fully automatic pipeline. We ensure our approach is fully automatic by ensuring every step in the pipeline can be performed without manual intervention.


For animating a head, most generative models use a 2D representation of the head \cite{siarohin2019first, zhao2022thin}. These models lack a structural representation of the human head, thus leading to artifacts and irregular shapes being formed, especially in extreme head poses. While there are 3D aware face-animation approaches for modeling \cite{blanz2023morphable, li2017learning} and animating human heads \cite{wang2021safa, yang2022face2face}, we could find only a few of the current de-identification approaches \cite{sun2018hybrid} which use them. To ensure a good structural quality, we use a 3D-aware animation approach.

\subsection{Important criteria}
We identify some of the major criteria needed to be fulfilled for a proper automatic de-identification pipeline for videos, and summarise them as follows:

\begin{enumerate}
    \item \textbf{Privacy:} Ensuring proper de-identification, without leaving any trace of the original identifiable part of the subject's face.
    \item \textbf{Realism:} The face in the output video should be realistic in terms of human likeness and visual fidelity. Realism should be achieved in 2 levels:
    \begin{enumerate}
        \item Image/Frame level: The face should resemble a valid human face in every frame and minimize artifacts. 
        \item Video level: The identity across all the faces should remain the same, i.e. there shouldn't be frame-wise flickering that can make the output look synthetic.
    \end{enumerate}
    \item \textbf{Preservation of non-identifiable attributes:} The output video must preserve behavioral features such as expression and non-verbal movements, which are non-identifiable.
    
    \item \textbf{Automated:} The pipeline should aim to minimise manual processing and decision-making to remove bias and subjectivity in the final results.

    \item \textbf{Diverse:} The pipeline should work for subjects from diverse groups of the population, where diversity could be in terms of skin color, age, gender, physical traits, and behavioral habits.
    
\end{enumerate}

To the best of our knowledge, we are the first to propose a solution that systematically addresses all these aspects.

\begin{figure}[t]

    \includegraphics[width=8.5cm]{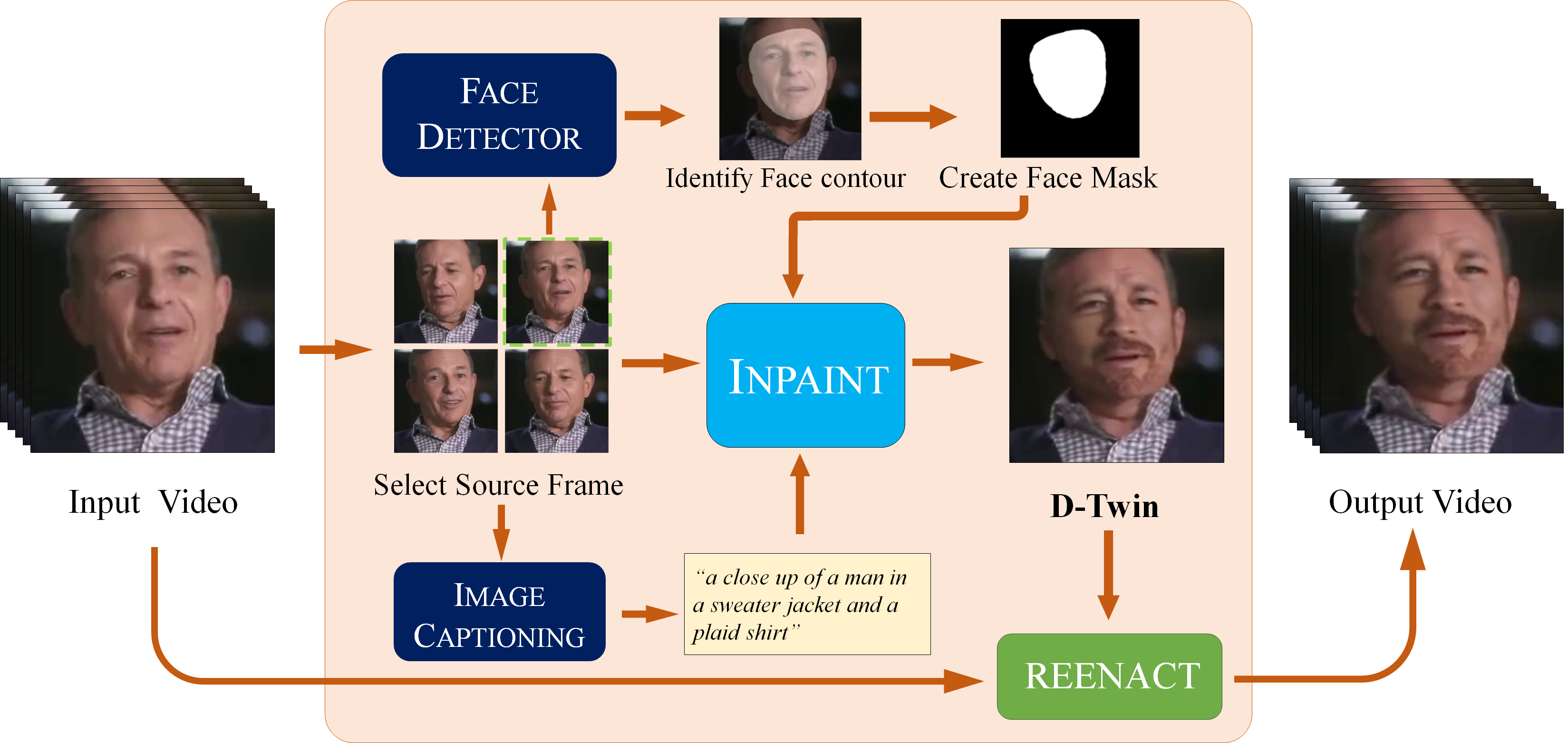}
    \caption{Pipeline of RID-Twin: From an input video $V_S$, we first extract a source image $I_S$. From here, the \textsc{FaceDetector} module detects a face to create a face mask $I_M$. The source image also goes to the \textsc{ImageCaptioning} module to generate a relevant caption $C_S$. The source image, mask, and caption go to the \textsc{Inpaint} module to generate the same image with a new identity $I_D$, i.e. our D-Twin. Finally, we re-enact this D-Twin image based on the motion of our input video using the \textsc{Re-enact} module, to get the output de-identified video $V_D$.}
    \label{fig:pipeline}
%
\end{figure}
\section{Proposed Methodology}
\label{sec:proposed}
In this section, we discuss our proposed approach RID-Twin. As shown in Fig.\ref{fig:pipeline}, our first goal is to get an actor who resembles the source subject in all aspects, except the identifiable features. The image of the subject we get after inpainting is our De-identified twin \textit{(D-Twin)} since it resembles the source subject in all aspects except for the identity (face). We then re-enact this synthetic actor with the motion of the source video. In contrast to the conventional approach of de-identifying frames of a video individually, we are essentially transferring a motion to a de-identified image (devoid of any identifiable trait). This unique approach of decoupling identity from motion ensures that intuitively there's no possible way to leak the original identity into the final video. While generating the D-Twin, features such as expression are lost. However, our re-enacting module takes care of this by transferring the expressions.
\break

\noindent\textbf{Overall pipeline:} Our proposed framework consists of four different modules: detecting the face to create a mask, captioning the face image, inpainting a masked-out image, and re-enacting (i.e. transferring a motion). For creating the D-Twin, given the input video $V_S$, we first find a source frame $I_S$ from the video. We chose the source frame to be one where the direction of the head pose of the subject is looking straight as much as possible. From this, the face detection module finds the face area and creates a corresponding mask $I_M\in (0,1)^{w\times h}$. This mask segments the face area in the image, instructing the inpainting module about the area that has to be generated. The source image $I_S$ also goes into the image captioning module, to generate a description $C_S$ for the subject in the image, which will also be used as a condition to the in-painting module. This description helps the inpainting module to preserve semantic details for the D-Twin. Compared to using a fixed set of attributes, a captioning module will be able to describe the images better and generalize to unseen settings. Then, we take the source image $I_S$, the generated mask image $I_M$, and the generated caption $C_S$, and send it to the inpainting module to generate $I_D \in \mathbb{R}^{H\times W\times3}$ which is our D-Twin. Finally, a face animation module takes in the source video $V_S$ and D-Twin image $ I_D $, and generates our final output video $ V_D\in\mathbb{R}^{H\times W\times3\times T}$ of the re-enacted D-Twin. This module can capture the pose, expression, and motion from the source video in a 3D-aware manner, thus preserving original structural and behavioral features. All the modules of the pipeline are captured in Algorithm \ref{alg:rid_twin}.
\break

\noindent\textbf{Face Detection:}
First, we need to identify a source frame, which contains the major portion of the face. We propose choosing the frame where the subject is closest to facing front, i.e. towards the camera. This ensures maximum visibility of the face area while generating a new face. To do this, we used the Mediapipe \cite{48292} for identifying the angle of rotation of the head. For every frame, we take the frame as the sum of the squares of the horizontal and vertical poses of the head. Then, we select the frame with the minimum value of this pose value as our source frame. From this frame, the \textsc{FaceDetection} module uses Mediapipe to extract the points of the face contour. This contour is used to build the face mask image $ I_M $ that segments out the face.
\break

\noindent\textbf{Image-captioning:}
To preserve the important semantic details of our source face, we generate a caption for the image, which will as a prompt to generate our D-Twin. The \textsc{ImageCaptioning} module captions the source image $ I_S$ using BLIP (Bootstrapping Language-Image Pre-training) \cite{https://doi.org/10.48550/arxiv.2201.12086}, a framework for vision-language understanding and generation task.
\break

\noindent\textbf{Inpainting:} 
For \textsc{Inpaint} module we use the Stable Diffusion \cite{rombach2022high} inpainting model. It is a state-of-the-art inpainting model that can generate photo-realistic output images from segmented-out input images and a prompt as guidance. Conditioned on the source frame, face mask, and the generated caption, this model outputs a generated image of D-Twin with the new face, i.e. our new identity. The generation is highly realistic, seamlessly aligning with the face structure, skin color, and also other features such as accessories (e.g. spectacles).
\break

\noindent\textbf{Face re-enactment:}
For the \textsc{Re-enact} module, we use Structure-Aware-Face-Animation (SAFA) \cite{wang2021safa} for animating the new image using the motion of the source video. SAFA preserves the pose, expression, and motion of the source video and transfers it to the generated one. It uses FLAME parameters, which is a parametric 3D Head model \cite{li2017learning}, and thus ensures structurally rich modeling of the subject's head while transferring the pose, expression, and motion.
 \begin{algorithm}[t]
 \caption{RID-Twin}
 \begin{algorithmic}[1]
 \renewcommand{\algorithmicrequire}{\textbf{Input:}}
 \renewcommand{\algorithmicensure}{\textbf{Output:}}
 \REQUIRE $ V_S\in\mathbb{R}^{H\times W\times3\times T} $ 
 \ENSURE  $ V_D\in\mathbb{R}^{H\times W\times3\times T} $ 
 \\ \textit{Initialisation}: pose: []

  \FOR {$i = 1 $ to $T$}
    \STATE $ pose[V_S^{(t = i)}] = pose_x(V_S^{(t = i)})^2 + pose_y(V_S^{(t = i)})^2$
  \ENDFOR
  \STATE $ I_S = V_S^{(t = i)} $ where $i \gets \arg \min_i pose[V_S] $
  \STATE $ I_M \gets \textsc{FaceDetector}(I_S)$ 
  \STATE $ C_S \gets \textsc{ImageCaptioner}(I_S)$ 
  \STATE $ I_D \gets \textsc{Inpaint}(I_S, I_M, C_S)$ 
  \STATE $ V_D \gets \textsc{Re-enact} (I_D, V_S)$ 
 \RETURN $V_D$ 
 \end{algorithmic}
 \label{alg:rid_twin}
 \end{algorithm}

\section{Experiments}
\label{sec:experiments}
\begin{figure}[ht]
    \includegraphics[width=8.5cm]{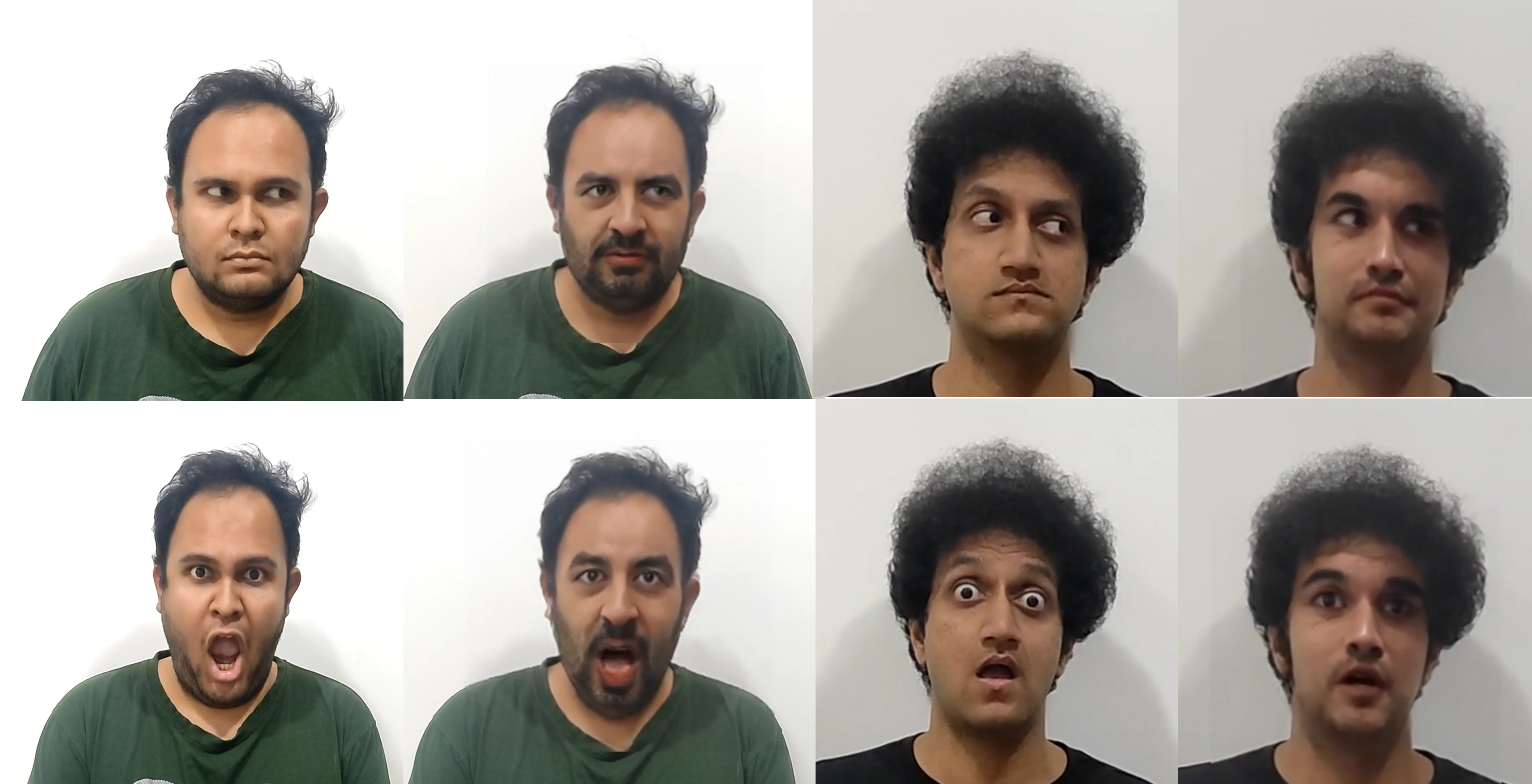}
    \setlength{\belowcaptionskip}{-17pt}
    \caption{RID-Twin on proposed custom dataset. In this figure, we see two examples of facial expression variation, the top row showing an instance of looking left, and the bottom row showing an instance of the expression 'shocked'. From left, we have frames from the first user's source video, the corresponding de-identified video, the second user's source video, and the corresponding de-identified video.}
    \label{fig:dataset_custom}
\end{figure}

We run RID-Twin for videos of talking human subjects and perform qualitative and quantitative evaluations of the results. We choose two datasets for our experiment: VoxCeleb2 \cite{nagrani2017voxceleb} and our custom dataset. VoxCeleb2 consists of videos of celebrities talking with varying head poses and expressions. We select the first part of the VoxCeleb2 test set, consisting of 118 identities. For our experiment, we select a single video for each of the 118 identities. Although the dataset is diverse in terms of human subjects, it lacks certain variations of behavioral features such as expressions and poses which are common in real-world scenarios. To see how well our approach performs in these situations, we propose a dataset with a set of motions and actions as an additional measurement of robustness. Our proposed custom dataset (Fig.\ref{fig:dataset_custom}) contains two users, each performing the following behaviors: \textit{Gaze variation, Expression variation, Speech along with head motion} and \textit{Rapid head pose changes}. 
These two datasets combined encompass a wide range of faces and behaviors for our evaluation.
\subsection{Evaluation} 
We evaluate the output videos of our approach on three crucial metrics:
\begin{enumerate}
    \item \textbf{De-identification level:} The identity of the source and D-Twin videos should be different.
    \item \textbf{Identity consistency:} The identity of D-Twin in a particular video should be consistent.
    \item \textbf{Expression preservation:} The expression of the source and D-Twin videos should be consistent.
\end{enumerate}

For a frame-by-frame evaluation of these metrics, we need to extract the frames of the source and their corresponding de-identified videos, and then find a measurement of the identity and the expression of the subjects on the frames. For each frame in the pair of videos (source and de-identified), we extract the face areas using MTCNN \cite{zhang2016joint}. We embed all the face frames on a 512-dimensional space of identity embedding using Face-Net \cite{schroff2015facenet}, and also on 16 dimensional space of expression embedding \cite{vemulapalli2019compact}. Based on these, we use cosine distances and Euclidean distances across all the frames of source and D-Twin videos to find the above-mentioned metrics.

\subsection{Results}
\begin{figure}[t]
    \includegraphics[width=8.5cm]{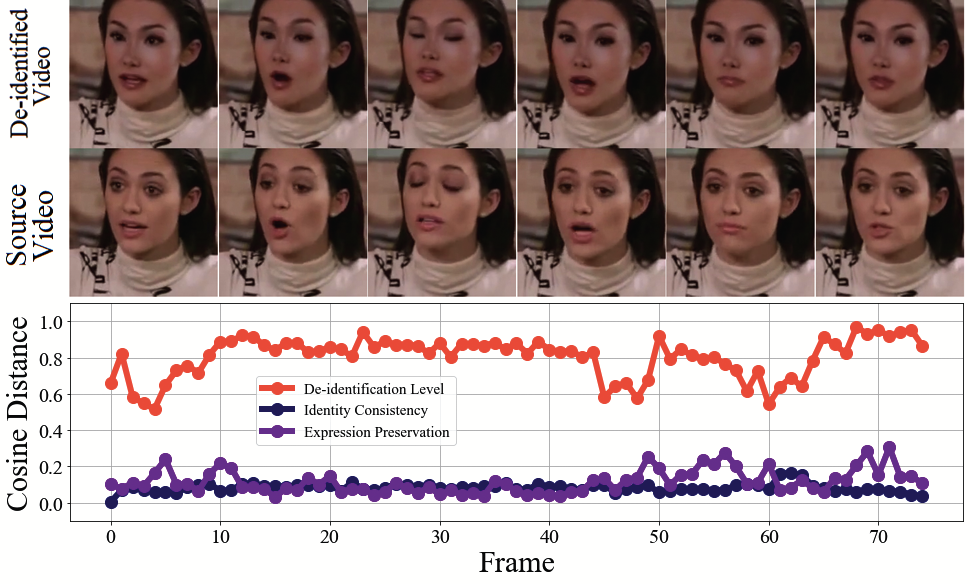}
    \caption{Preservation of expression and pose across frames: \textit{(From top)} De-identified video, Source Video, and Plots of Evaluation metrics: De-identification Level, Identity Consistency, Expression Preservation using Cosine Distance}
    \label{fig:output_timeline_result}
\end{figure}
We have visualized some of the outputs of generated D-Twins in Fig.\ref{fig:output_dtwins} \footnote{https://github.com/AnirbanMukherjeeXD/RID-Twin}
We see that the generated D-Twins preserve the outer facial regions from the source video, and generate new faces with completely different identities. We also see in the videos that the expressions in the de-identified output video are the same as those of the source video (Fig. \ref{fig:output_timeline_result}).
To quantify our results, we use the metrics of deidentification level, identity consistency, and expression preservation. The de-identification level is calculated as the distance between the identity embeddings of the frames of the source video and the de-identified video. The identity consistency is calculated as the distance between the identity embeddings of the current frame and that of the first frame of the de-identified video.  The expression preservation is calculated as the distance between the expression embedding of the frames of the source video and the de-identified video. Visualizing these metrics for every frame gives us a detailed evaluation of the performance of our model on the de-identified videos. We show a plot of these metrics along with the frames of the de-identified video in Fig. \ref{fig:output_timeline_result}. For an average result of a video, we find the mean and variances of these metrics across all the frames. To summarize our results across the whole dataset, we find the means of these values across all the converted videos. We can see from these values that the identity metrics are on average different for source and de-identified videos (De-identification level) denoted by a high distance value, and the same for the frames across a de-identified video (Identity consistency) denoted by a low distance value. We also see that expression metrics are on average the same for the source and de-identified videos (Expression preservation), denoted by a low distance value. We propose that these metrics are an appropriate way of quantifying these crucial aspects for the task of face de-identification.
\section{DISCUSSION}
\label{sec:discussion}
\begin{table}[t]
\caption{Quantitative results on VoxCeleb2 dataset}
\label{lit_survey}
\begin{center}
\begin{tabular}{|l||c|c|}
\hline
\textbf{Criteria} &  \textbf{Means}&  \textbf{Variances}\\
\hline
\multicolumn{3}{|l|}{\textit{Cosine Distances}}\\
\hline
De-identification Level $\uparrow$ & 0.745 & 007 \\
Identity Consistency $\downarrow$ & 0.170 & 006\\
Expression Preservation $\downarrow$ & 0.100 & 002\\
\hline
\multicolumn{3}{|l|}{\textit{Euclidean Distances}}\\
\hline
De-identification Level $\uparrow$ & 1.207 & 0.005\\
Identity Consistency $\downarrow$ & 0.549 & 0.017\\
Expression Preservation $\downarrow$ & 0.404 & 0.009\\
\hline
\end{tabular}
\end{center}
\end{table}
In our work, we identify five crucial aspects of video de-identification tasks: privacy, realism, non-identifiable feature preservation, automation, and diversity. We show how RID-Twin is capable of addressing these points. To ensure complete privacy, we chose to re-enact a de-identified twin based on the source subject. We do not perform any frame-wise update for the videos, and the only transfer that happens is the motion-related details, such as key points from the source video to the D-Twin. For realism of the de-identified images, we use the current state-of-the-art generative model, i.e. Stable Diffusion, for generating photorealistic faces. To ensure realism in terms of movements, we use a motion transfer instead of generating the face in a per-frame manner, to ensure temporal inconsistency. Our chosen \textit{3D-aware} motion transfer model has the capability of modeling the structural intricacies of the human face, thus leading to a better transfer of motion, compared to purely 2D-based approaches. This also preserves non-identifiable behavioral features such as expressions and poses. This could be useful for downstream tasks such as automatic clinical analysis, or manual inspection by humans. To make our proposed pipeline fully automatic, we include key steps such as automated selection of the source frame to be inpainted and preservation of appearance details of the source image using text captions. End-to-end automation reduces manual intervention and removes any possibility of human bias in the process. Finally, we ensure our pipeline is diverse in terms of attributes such as age, skin color, and physical features. Since inpainting is done based on the background of the human head, the pipeline takes care of variations in faces. Using a prompt ensures some other details such as accessories and spectacles exist on the inpainted faces. Moreover, since Stable Diffusion has been trained on a huge variety of datasets, we rely on it to be able to produce diverse faces. Through the metrics we use, we can ensure whether the source video has been accurately de-identified, while the expressions have been preserved. Our focus on these crucial aspects ensures a trustworthy and effective pipeline for automatic face de-identification for videos. 

\noindent\textbf{Limitations and future work:} The performance of the proposed approach depends on the corresponding models that we have taken in our pipeline. While our chosen state-of-the-art models produce highly realistic outputs, there still might be erroneous cases, which will be evident from our metrics. Given the offline nature of our pipeline, in such cases, the user can run our pipeline multiple times by varying the parameters of the module to increase the chances of improved outputs. In the future, as these modules become more robust to varying cases, the failure rate will drop. Our goal here is not to present this work as a one-stop solution but as a framework that can be improved and fine-tuned on a case-to-case basis. Our work also depends on the assumption that we can identify at least one frame where the user looks towards the camera. However, this may not always be the case in practice, leading to some uncertainties for in-the-wild tasks. Although it may not lead to aesthetically pleasing outputs, our formulation covers this issue in general. Currently, our approach works on a single subject. In the future, it can be extended to work on multiple subjects in a video. Another interesting direction could be to make our approach real-time. Using a live motion transfer approach in the \textsc{Re-enact} module could enable the pipeline to function in an online mode and be useful in live face de-identification settings. While we used identity and expression embeddings as evaluation metrics, other aspects such as action units, gaze, lip sync, and other behavioral attributes can be used for further experimentation on domain-specific downstream applications such as healthcare.

\section{CONCLUSION}
\label{sec:conclusion}
In this work, we propose RID-Twin: a novel way of building an end-to-end automatic pipeline for performing face de-identification for videos. We look at the crucial persisting challenges in de-identification and show how the proposed approach systematically addresses them by utilizing state-of-the-art generative models and decoupling motion from identity. Furthermore, we discuss how to evaluate RID-Twin to ensure that the desired criteria are met. We demonstrate the performance of the pipeline on the VoxCeleb2 dataset, and also the proposed custom dataset.  Our work is a novel step towards solving the problem of face de-identification and could serve as a benchmark for future applications.




\bibliographystyle{IEEEbib}
\bibliography{refs}

\end{document}